\documentclass{article}
\usepackage{iclr2021_workshop,times}

\usepackage{amsmath,amsfonts,bm}

\def\eqref#1{equation~\ref{#1}}

\def\1{\bm{1}}

\DeclareMathAlphabet{\mathsfit}{\encodingdefault}{\sfdefault}{m}{sl}
\SetMathAlphabet{\mathsfit}{bold}{\encodingdefault}{\sfdefault}{bx}{n}

\usepackage{hyperref}
\usepackage{url}
\usepackage[ruled, vlined]{algorithm2e}
\usepackage{graphicx,subcaption,ctable,tabularx}

\title{On Linear Interpolation in the Latent Space of Deep Generative Models}

\author{Mike Yan Michelis \\
Department of Computer Science\\
Technical University of Munich, Germany\\
\texttt{mike.michelis@tum.de} \\
\And
Quentin Becker \\
Geometric Computing Laboratory \\
{\'E}cole Polytechnique F{\'e}d{\'e}rale de Lausanne, Switzerland \\
\texttt{quentin.becker@epfl.ch} \\
}

\iclrfinalcopy 
\begin{document}

\maketitle

\begin{abstract}


The underlying geometrical structure of the latent space in deep generative models is in most cases not Euclidean, which may lead to biases when comparing interpolation capabilities of two models. Smoothness and plausibility of linear interpolations in latent space are associated with the quality of the underlying generative model. In this paper, we show that not all such interpolations are comparable as they can deviate arbitrarily from the shortest interpolation curve given by the geodesic. This deviation is revealed by computing curve lengths with the pull-back metric of the generative model, finding shorter curves than the straight line between endpoints, and measuring a non-zero relative length improvement on this straight line. This leads to a strategy to compare linear interpolations across two generative models. We also show the effect and importance of choosing an appropriate output space for computing shorter curves. For this computation we derive an extension of the pull-back metric. Code available at: \url{https://github.com/mmichelis/GenerativeLatentSpace}


\end{abstract}


\section{Introduction}


Generative models trained in frameworks such as Generative Adversarial Networks (GAN) \citep{goodfellow2014generative} or Variational Autoencoder (VAE) \citep{kingma2013vae} have achieved exciting results in computer vision \citep{karras2020analyzing}. In its simplest form, the trained generative model $g$ maps a latent space $\mathcal{Z}$ to some output space $\mathcal{X}$. Vectors populating $\mathcal{Z}$ are sampled according to an arbitrary and fixed latent distribution: $\mathbf{z}\sim\mathbb{P}_Z$, often chosen to be Gaussian. Training, in generative modeling, consists of approximating a target distribution $\mathbb{P}_T$ with the output distribution $g(\mathbf{z})\sim\mathbb{P}_G$.

$\mathbb{P}_Z$ is distorted by the generator in order to fit the modal characteristics of $\mathbb{P}_T$, creating a highly nonlinear mapping as a result. Hence, navigating in latent space may incur very dissimilar transformations to the output depending on the starting point and the direction picked. Understanding this latent space has been attempted in the form of e.g., finding ``meaningful directions'' \citep{voynov2020unsupervised, shen2020interpreting}.

Following \citet{arvanitidis2018latentspace}, we know that this latent space cannot be regarded as Euclidean without further investigation. Instead, it is equipped with a Riemannian induced metric $\textbf{M} = \textbf{J}^T \textbf{J}$, also called ``pull-back metric'' \citep{daouda2020geometry, riemmaniangeometry}, where $\textbf{J} = \frac{\partial g}{\partial \boldsymbol{z}}$ is the Jacobian of the generator. The choice of the output space greatly affects the induced metric \citep{laine2018feature}, a choice that we explore in Section \ref{sec:feat}.


We intend to use the previously defined Jacobian to evaluate the average quality of linear interpolations over the whole latent space in the following manner: First, we define the Riemannian curve length, then we sample many shorter curves in latent space and compute the relative length improvement of each quasi-geodesic over the corresponding straight line. The average and standard deviation of these relative improvements are then used as evidence for supporting the discrepancy in linear interpolation quality. Based on these observations, we propose a strategy for comparing interpolations across generative models.

\section{Method}

\subsection{Shorter Curve}

As the generator is highly nonlinear, its associated pull-back metric isn't constant and equal to the identity over the latent space. Hence a straight line is unlikely to be a geodesic, and we can find curves with shorter length that connect two points in latent space. This, in turn, enables a better evaluation of the distance between points. To find shorter curves/geodesics in the Riemannian manifold given by the pull-back metric, we first define the length of a curve $\boldsymbol{\gamma}(t) =: \boldsymbol{\gamma}_t$ (defined for $t \in [0,1]$ and using shorthand notation $\textbf{J}_{\gamma} := \textbf{J}(\gamma(t))$):
\begin{align}
    \label{eq:len}
    \begin{split}
    \text{Len}(\gamma) = \int_0^1 \| \dot{g}(\boldsymbol{\gamma}_t) \| dt = \int_0^1 \| \textbf{J}_{\gamma} \dot{\boldsymbol{\gamma}}_t \| dt 
    = \int_0^1 \sqrt{ \dot{\boldsymbol{\gamma}}_t^T  \textbf{J}_{\gamma}^T \textbf{J}_{\gamma} \dot{\boldsymbol{\gamma}}_t } dt
    \end{split}
\end{align}
Next we choose a method of representing/implementing the curve. Existing methods include e.g., \citet{DBLP:conf/aistats/ArvanitidisHHS19} using Gaussian Processes, \citet{yang_geodesicClustering} using quadratic functions, or \citet{laine2018feature} using a discrete array of points. We opted for a continuous curve with analytical derivatives, where control points only change behavior locally: B-splines. As we need at most second order derivatives, we chose cubic B-splines (see implementation details in Appendix \ref{app:bspline}). 


With this curve implementation, we can either directly minimize $\text{Len}(\gamma)$ via optimization, or solve the geodesic Ordinary Differential Equation (ODE) defined and derived in \citet{arvanitidis2018latentspace} for finding the minimizer $\gamma$ of Equation \ref{eq:len}. The ODE requires a second derivative i.e., the Hessian, which is computationally much more expensive, and furthermore did not consistently find shorter curves than the direct length minimization approach in our experiment. The ODE approach does, however, provide a measure of convergence, i.e. how close the solution is to the geodesic. In the end, we chose not to use the ODE, and instead minimize Equation \ref{eq:len} for finding shorter curves in latent space by optimizing the control points of the cubic B-spline using gradient descent. To improve the convexity of $\text{Len}(\gamma)$ and have a better behaved optimization, we instead minimize $\int_0^1 \| \dot{g}(\boldsymbol{\gamma}_t) \|^2 dt$ i.e., the Path Energy, which does not change the minimizer of the former functional.


We initialize the cubic B-spline with a straight line with fixed start and end points plus two variable control points in between. Whenever the curve length plateaus, we add a new control point and resume optimization. Termination criteria are maximal node count and number of optimization steps. The drawback of this method is that we cannot claim that the result is a geodesic, it is simply a shorter curve than the straight line. Consequently, its length provides a more accurate ``distance'' between points than measuring the straight line's Riemannian length.


\subsection{Jacobian}
\label{sec:feat}



The pull-back metric depends solely on the Jacobian of the generator, which requires special care when being computed. For a deterministic generator, it is enough to backpropagate the derivatives from the output of the generator to the inputs. In the case of a stochastic generator/decoder as in VAE, we use the expected value of the induced metric, combining the Jacobian for the mean and standard deviation of the outputs \citep{arvanitidis2018latentspace}:
\begin{align}
    {\textbf{M}}_z := \overline{\textbf{M}}_z = \left( \textbf{J}^{\mu}_z \right)^T \textbf{J}^{\mu}_z + \left( \textbf{J}^{\sigma}_z \right)^T \textbf{J}^{\sigma}_z
\end{align}

An open question is whether the output space $\mathcal{X}$ of the generator (which is often an image) is meaningful for the metric computation, hence it is worth investigating what the effect is of ``feature mappings'' $f: \mathcal{X} \to \mathcal{F}$ on top of the generator. Two options we implemented were the logistic regression output and activations of a VGG-19 network. For the former, a clear and intuitive effect can be observed when tested for MNIST digits: geodesics pass through as few digit clusters as possible in latent space (see Appendix \ref{app:seq}). VGG activations are assumed to structure the latent space in a perceptually meaningful way: \citet{laine2018feature} investigates this claim qualitatively, while \citet{moor2020challenging} does so quantitatively.

For further (deterministic) mappings on top of a deterministic generator, we simply multiply the Jacobians to compute the overall induced metric:
\begin{align}
    \textbf{M}_{FZ} := \textbf{J}^T_{XZ} \textbf{M}_{FX} \textbf{J}_{XZ}
\end{align}
Here $\textbf{J}_{XZ}$ stands for the Jacobian of the generator $g: \mathcal{Z} \to \mathcal{X}$, and $\textbf{J}_{FX}$ would be for the feature mapping $f: \mathcal{X} \to \mathcal{F}$, with $\textbf{M}_{FX} := \textbf{J}^T_{FX} \textbf{J}_{FX}$. In the case of a stochastic generator (under assumption of diagonal covariances):
\begin{align}
    \textbf{M}_{FZ} := \overline{\textbf{M}}_{FZ} = \left( \textbf{J}^{\mu}_{XZ} \right)^T \textbf{M}_{FX} \textbf{J}^{\mu}_{XZ} + \left( \textbf{J}^{\sigma}_{XZ} \right)^T \hat{\textbf{M}}_{FX} \textbf{J}^{\sigma}_{XZ}
    \label{eq:feature}
\end{align}

Here $\hat{\textbf{M}}_{FX}$ keeps just the diagonal entries of $\textbf{M}_{FX}$, i.e., all off-diagonals are set to 0. We only need the diagonals for variance as we assume diagonal covariances for the normal distribution in output space (e.g. VAE). The exact derivation of this result can be found in Appendix \ref{app:feature}.

\subsection{Evaluation of Linear-to-Geodesic Deviation}
\label{sec:mc}

To quantify how much the linear interpolations deviate from the geodesics on average, we define and measure the expected worst-case relative improvement of the Riemannian length between pairs of points by sampling over the whole latent space. We can sample a starting point from the latent distribution $\mathbb{P}_Z$, and move in the direction of the eigenvector with the largest eigenvalue of the pull-back metric at that point; a direction we call maximal eigenvector for short. For a given step-size, we now have a start and end point, and can compute the relative improvement of a shorter curve compared to the Riemannian length of the straight line connecting both points. Lastly, we take an average of this value over all the samples in latent space. The process can be described as in Algorithm \ref{alg:improv} in Appendix \ref{app:mc}.

\begin{figure}[htb]
    \centering
    \begin{subfigure}{0.49\textwidth}
        \centering
        \includegraphics[width=\textwidth]{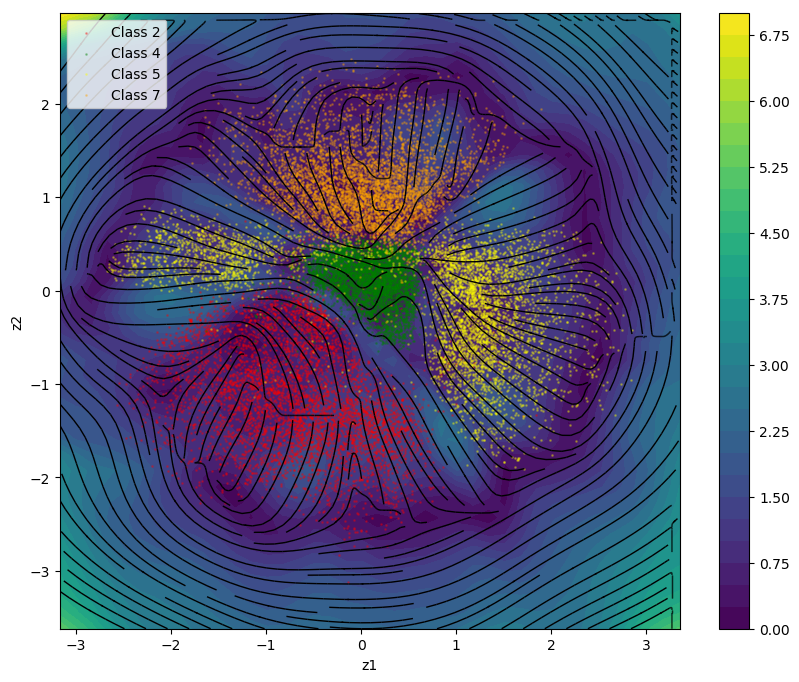}
    \end{subfigure}
    \hfill
    \begin{subfigure}{0.49\textwidth}  
        \centering 
        \includegraphics[width=\textwidth]{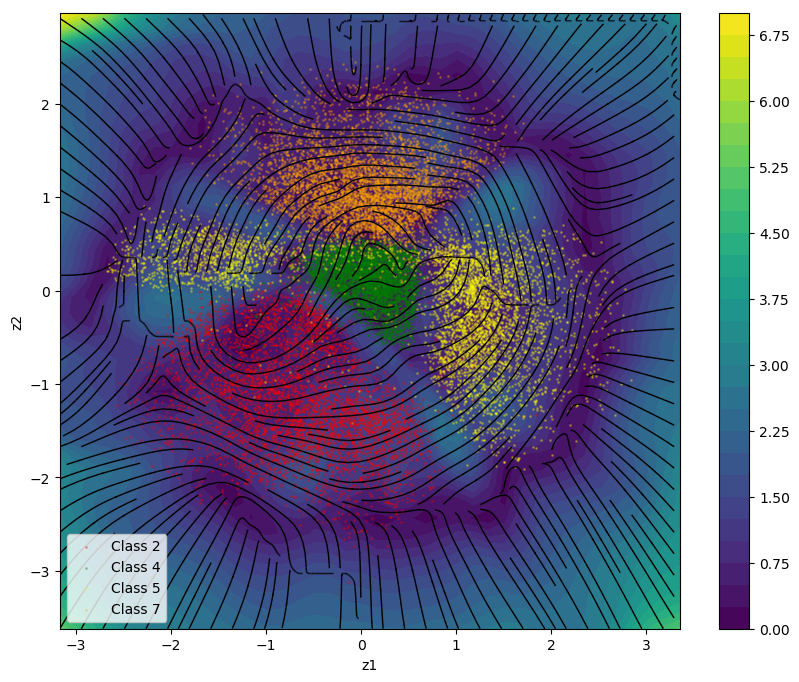}
    \end{subfigure}
\caption{The latent space of this VAE is 2D. In the background the logarithmic condition number i.e., ratio of largest to smallest eigenvalue is plotted, together with clusters of encoded MNIST test data (digits 2,4,5,7). In the foreground we see streamlines following the minimal eigenvectors i.e., eigenvectors of smallest eigenvalue of pull-back metric at every point on the left, and maximal eigenvectors on the right. We used the improved VAE variance estimate of \citet{arvanitidis2018latentspace}.}
\label{fig:streamline}
\end{figure}

The result on a VAE can also be found in Appendix \ref{app:mc}. We follow the maximal eigenvectors at a given point in the latent space\footnote{We aligned the maximal eigenvectors such that it always points towards the origin, this prevents the end point to lie in regions where the generator was not trained for} to induce maximal change in output, which computes the ``worst-case'' and enables us to obtain an upper bound on how much the worst straight lines can be improved. We found that the metric is highly anisotropic (see the large condition numbers in Figure \ref{fig:streamline}), and while \citet{wang2021a} observed that the maximal eigenvectors are similar at different positions in the latent space, we remark that for our experiments they were not homogeneous, but varied in direction throughout latent space (see Figure \ref{fig:streamline}).

\newpage

\section{Discussion}
\label{sec:disc}

\begin{figure}[htb]
\begin{center}
\begin{subfigure}{0.49\textwidth}
    \centering
    \includegraphics[width=\textwidth]{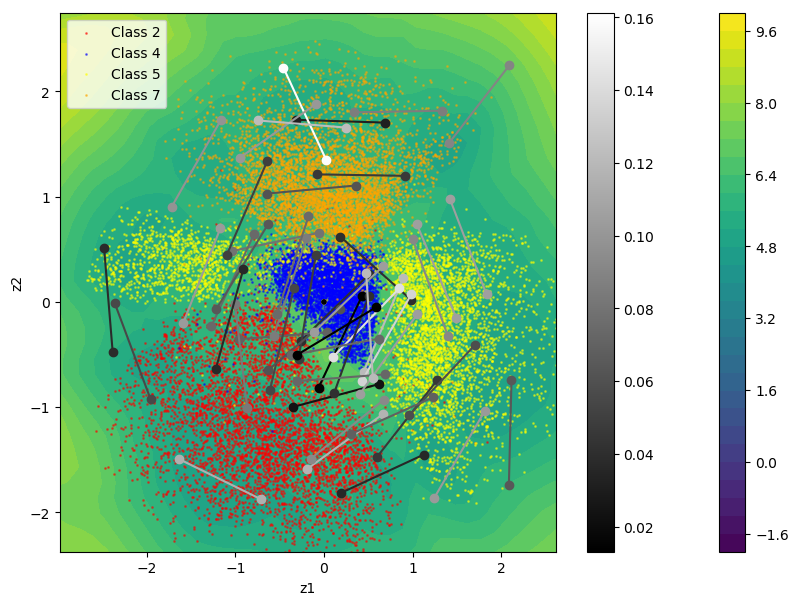}
\end{subfigure}
\hfill
\begin{subfigure}{0.49\textwidth}  
    \centering 
    \includegraphics[width=\textwidth]{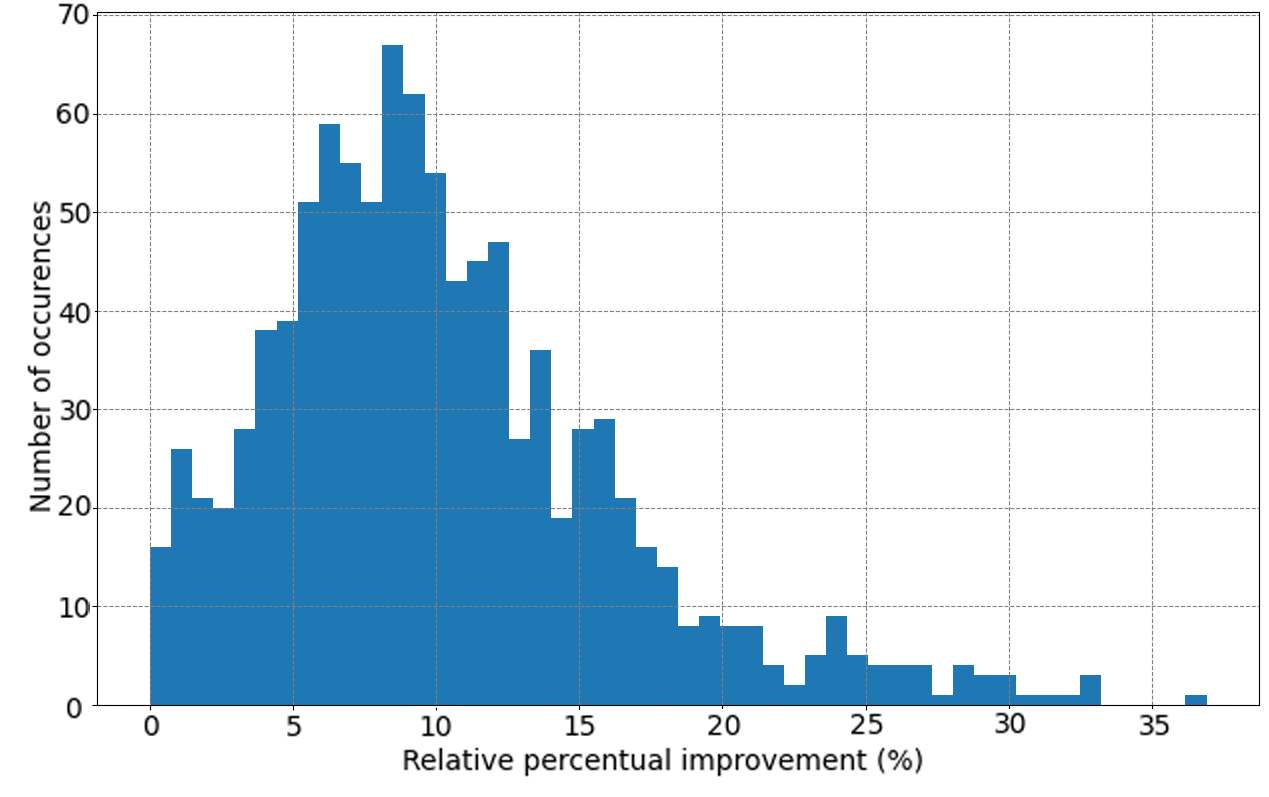}
\end{subfigure}
    
\end{center}
\caption{The latent space of this VAE is 2D. \textit{Left}: In the background $\log \sqrt{\det{\textbf{M}_z}}$ is plotted in color, together with clusters of encoded MNIST test data (digits 2,4,5,7). In the foreground we see straight lines connecting sampled pairs of points as described by Algorithm \ref{alg:improv}, where the brightness indicates how much the curve length can be shortened (0.14 indicates that the shorter curve is 14\% shorter than the straight line Riemannian length). \textit{Right}: Histogram of the distribution of relative length improvements using 1000 samples. A larger version can be found in Appendix \ref{app:mc}.}
\label{fig:MC}
\end{figure}

Visually smooth linear interpolations are often interpreted as a marker of the generative model's performance. Yet this criterion greatly depends on the place where the linear interpolation is performed in latent space, as it can be arbitrarily far off the geodesic (see Figure \ref{fig:MC}). As a result, we believe considering the whole latent space (e.g. through Monte Carlo sampling) was more statistically meaningful than evaluating several hand-picked interpolations.

From Figure \ref{fig:MC} we observe that the ``expected worst-case relative improvement'' (in this case $10.20\%$) is non-zero, which shows that the deviation of the straight line from the geodesic is significant enough to take into consideration. Additionally, the standard deviation of the histogram shows how we should be careful in choosing interpolations, as the relative improvement could be vastly different at different locations in latent space.

We present an example strategy of choosing latent linear interpolations in order to compare the interpolation quality of two generative models. We start by randomly sampling pairs of input data points from the test dataset, which we then encode into both latent spaces. In case no encoder is available for the generative model, one can search for closest matching latent points that generate the input samples, as shown in \citep{lei2019inverting} for instance. By finding a shorter curve connecting the pairs of points, we can compute relative length improvements for each of the two generators. Then, we choose the pair of points that has the most similar relative improvement, which can be seen as them being located in a similar metric neighborhood (see results in Appendix \ref{app:linint}). This choice rules out the comparisons between the best linear interpolation of one model to the worst of the other.

\section{Conclusion}


Accounting for the non-Euclidean nature of the latent space of generative models, we present a geometry-aware method for indicating the comparability of latent linear interpolations. Through random sampling, an average measure can be obtained over the whole latent space. A limitation of our current implementation is that we do not find global length minimizers/geodesics. Our computation for shorter curves could also be improved (as in \citet{DBLP:conf/aistats/ArvanitidisHHS19}), mainly due to the expensive computation of the Jacobian. We showed that the latent space geometry cannot be disregarded while evaluating generative models, and we believe geometry may further help us understand deep learning in the future.



\newpage

\bibliographystyle{iclr2021_workshop}
\bibliography{iclr2021_workshop}

\newpage

\appendix

\section{Cubic B-Spline}
\label{app:bspline}

The implementation of the cubic B-spline is as follows:
\begin{align*}
    N_{i,1}(t) &=  \begin{cases} 1 & \text{for}\ t_i \leq t < t_{i+1} \\ 0 & \text{otherwise} \end{cases} 
    \\
    N_{i,k}(t) &=  \frac{t - t_i}{t_{i+k-1} - t_i} N_{i,k-1}(t) + \frac{t_{i+k} - t}{t_{i+k} - t_{i+1}} N_{i+1,k-1}(t)
    \\
    \boldsymbol{C}(t) &= \sum_{i=0}^n N_{i,4}(t) \boldsymbol{P}_i
    \\
    \boldsymbol{T} &= (\underbrace{0,0,0,0}_{4}, \underbrace{knots(]0,1[)}_{n+1-4}, \underbrace{1,1,1,1}_{4})
\end{align*}

Here $N_{i,k}$ are the basis functions, $\boldsymbol{P}_i$ are the control points and $\boldsymbol{T}$ is the knot vector. We want end-point interpolation, hence we have the knot multiplicities above for start and end point. This b-spline has $n+1$ control points and order 4. Accordingly the first derivative is defined as:
\begin{align*}
    \frac{\partial \boldsymbol{C}(t)}{\partial t} = 3 \sum_{i=0}^{n-1} N_{i+1,3}(t) \frac{\boldsymbol{P}_{i+1} - \boldsymbol{P}_{i}}{T_{i+4} - T_{i+1}}
\end{align*}

The knot vector consists of a middle part with $n-3$ elements, which is adjusted everytime a new control point is added. We wish to add new control points in such a way that the old curve geometry is preserved. The implementation (choosing where to put new knots first) is as follows: we find the largest knot interval, and place a new knot in the exact middle of it. Knowing the new knot, we insert a new control point accordingly by adjusting 2 old ones ($i \in [j-2, j]$ where $j$ is the knot index of the new knot) such that curve geometry is kept the same:
\begin{align*}
    \boldsymbol{P}^{new}_i \longleftarrow &(1-a_i) \boldsymbol{P}^{old}_{i-1} + a_i \boldsymbol{P}^{old}_i
    \\
    a_i = &\frac{t_{new} - t_i}{t_{i+k-1} - t_i}
\end{align*}
We update from largest $i$ to smaller ones, so we always use the old values and do all operations in-place. $t_i$ are the knots. The point $\boldsymbol{P}^{new}_j$ is not updated in the old control point list, but will instead be our new control point that we insert at position $j$ in the list.

\section{Effect of logistic regression feature mapping}
\label{app:seq}

\begin{figure}[htb]
\begin{center}
\includegraphics[width=0.9\textwidth]{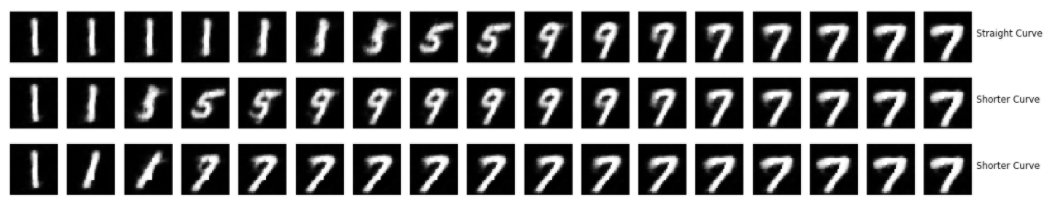}
\end{center}
\caption{Interpolating between two fixed points in latent space of a VAE trained on MNIST digits. Top sequence is along the straight line, middle is shorter distance as defined by the pull-back metric of the generator, and bottom is according to metric with a logistic regression feature mapping on top of generator.}
\label{fig:seq}
\end{figure}

As the logistic regression maps an image to probabilities of every class, finding a geodesic in such a pull-back metric is equivalent to travelling fastest between classes, i.e. traversing as few classes as possible. The effect can be observed in Figure \ref{fig:seq}, where the middle sequence is without any feature mapping, and the bottom one has the logistic regression, skipping over several digits in between.

\section{Monte Carlo Relative Improvements}
\label{app:mc}

The algorithm described in Section \ref{sec:mc} can be formalized as:

\begin{algorithm}
    \SetAlgoLined
    \SetKwInput{KwInput}{Input}
    \KwInput{Step size $\alpha$}
    \KwResult{Mean of relative length improvements}
    
    \texttt{\\}
    $L \gets$ empty list\;
     
    \While{maximal samples not reached}{
        $\boldsymbol{x}_A \gets$ sample latent vector\;
        $\boldsymbol{v}_{max} \gets$ eigenvector of metric with largest eigenvalue at $\boldsymbol{x}_A$\;
        $\boldsymbol{x}_B \gets \boldsymbol{x}_A + \alpha \boldsymbol{v}_{max}$\;
        \texttt{\\}
            
        $d_s \gets$ compute Riemannian length of straight line\;
        $d_c \gets$ find shorter length\;
        $relative\_improvement \gets \frac{d_s - d_c}{d_s}$\;
        Append $relative\_improvement$ to $L$\;
    }
    
    \texttt{\\}
    \Return{mean of $L$}\;

\caption{Expected worst-case relative improvement}
\label{alg:improv}
\end{algorithm}

\begin{figure}[htb]
\begin{center}
\includegraphics[width=1.\textwidth]{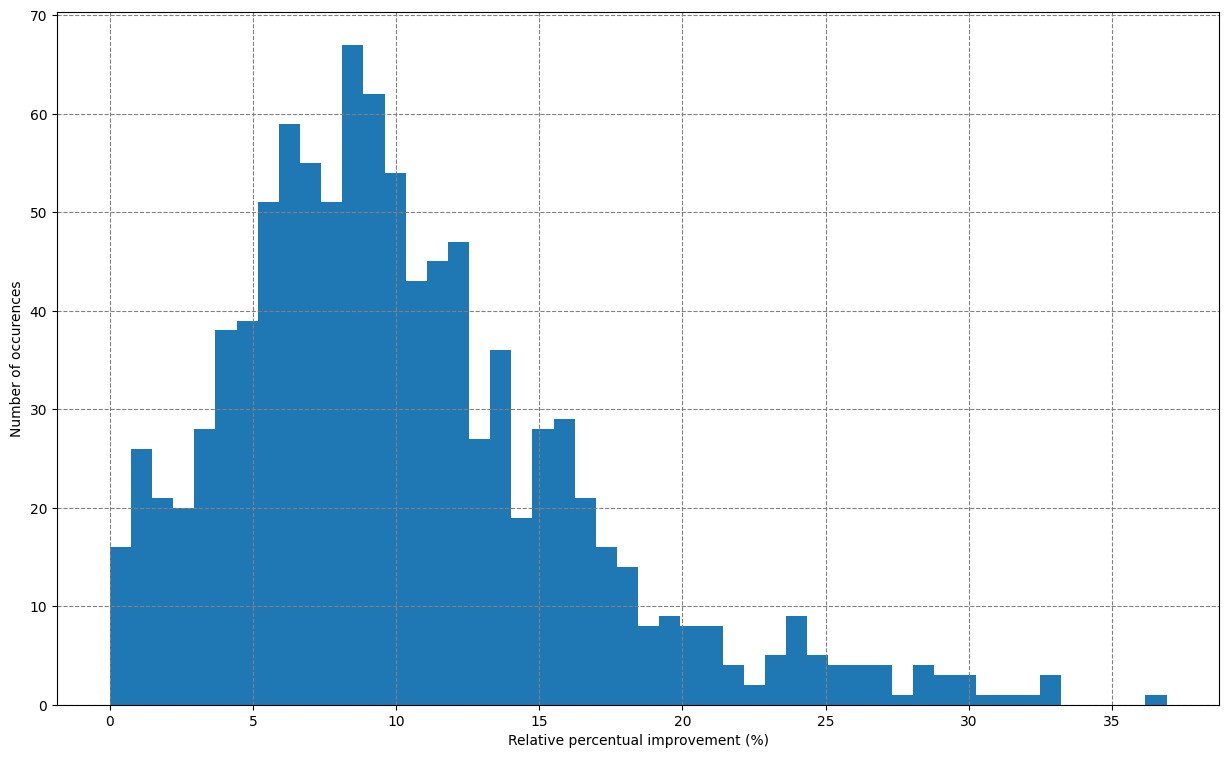}
\end{center}
\caption{Histogram of the distribution of relative length improvements for a VAE latent space using 1000 samples according to Algorithm \ref{alg:improv}.}
\label{fig:relimprov}
\end{figure}

For a concrete example, we ran it on a trained VAE for MNIST digits. After acquiring a list of relative improvements, we plot the frequency of occurrence in certain intervals, to get a good idea of how much the relative improvement varies in magnitude, see Figure \ref{fig:relimprov}.

\section{Comparability of Latent Linear Interpolations: Example Strategy}
\label{app:linint}

\begin{figure}[htb]
\begin{center}
\includegraphics[width=1.\textwidth]{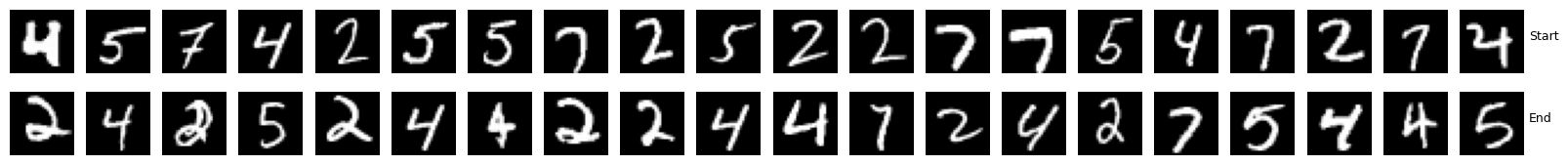}
\end{center}
\caption{Start and end points of 20 input space samples.}
\label{fig:samplesinput}
\end{figure}

\begin{figure}[htb]
\begin{center}
\includegraphics[width=0.9\textwidth]{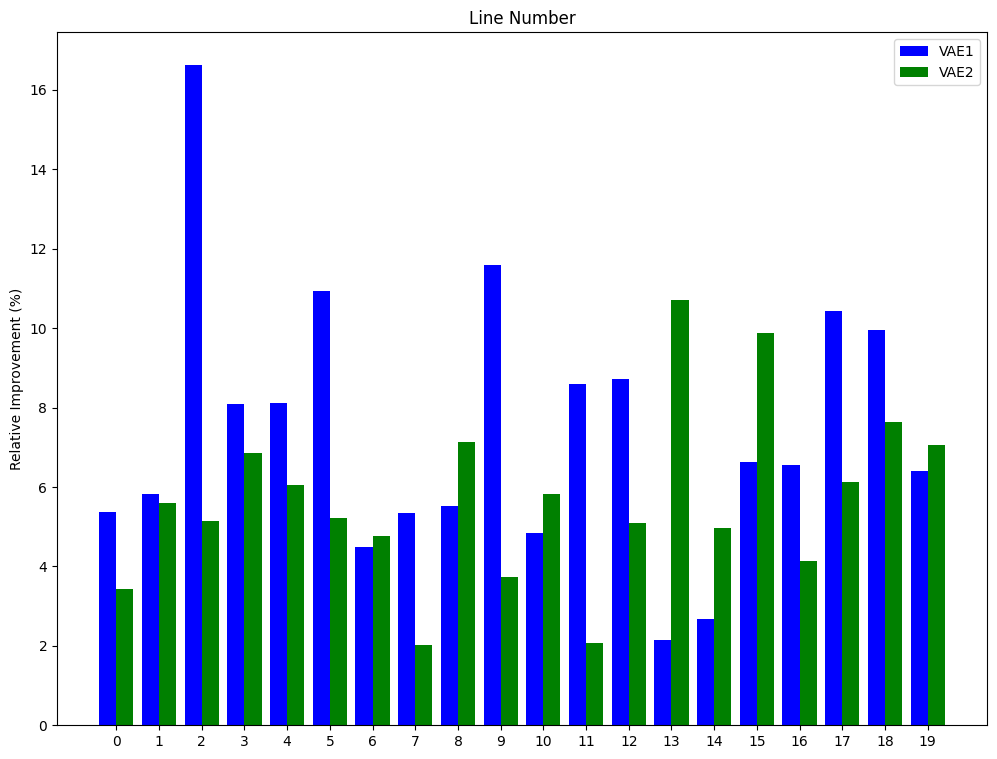}
\end{center}
\caption{Bar plot of relative improvements for 20 random sampled latent linear interpolation for two VAEs.}
\label{fig:relimprovlin}
\end{figure}

In the following we trained two VAEs with different architecture (both 2D latent space) on MNIST digits 2,4,5 and 7, and compared their interpolation capability as described in Section \ref{sec:disc}. VAE1 has 16 times more trainable parameters than VAE2. We random sampled 20 pairs of start and end points in input space, which can be seen in Figure \ref{fig:samplesinput}. We then found the relative improvements that are possible for each pair using both models, which can be seen in Figure \ref{fig:relimprovlin}.

\hfill \vspace{1em}

\begin{figure}[htb]
    \centering
    \begin{subfigure}{0.49\textwidth}
        \centering
        \includegraphics[width=\textwidth]{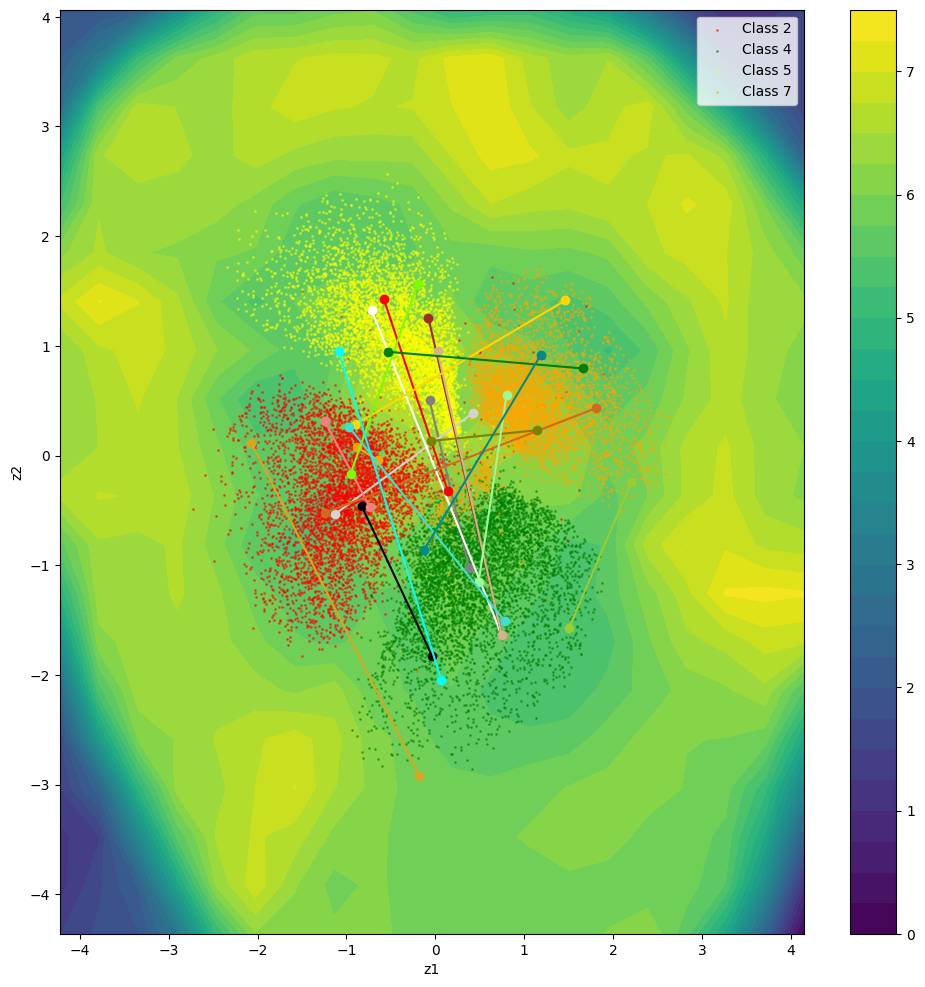}
    \end{subfigure}
    \hfill
    \begin{subfigure}{0.49\textwidth}  
        \centering 
        \includegraphics[width=\textwidth]{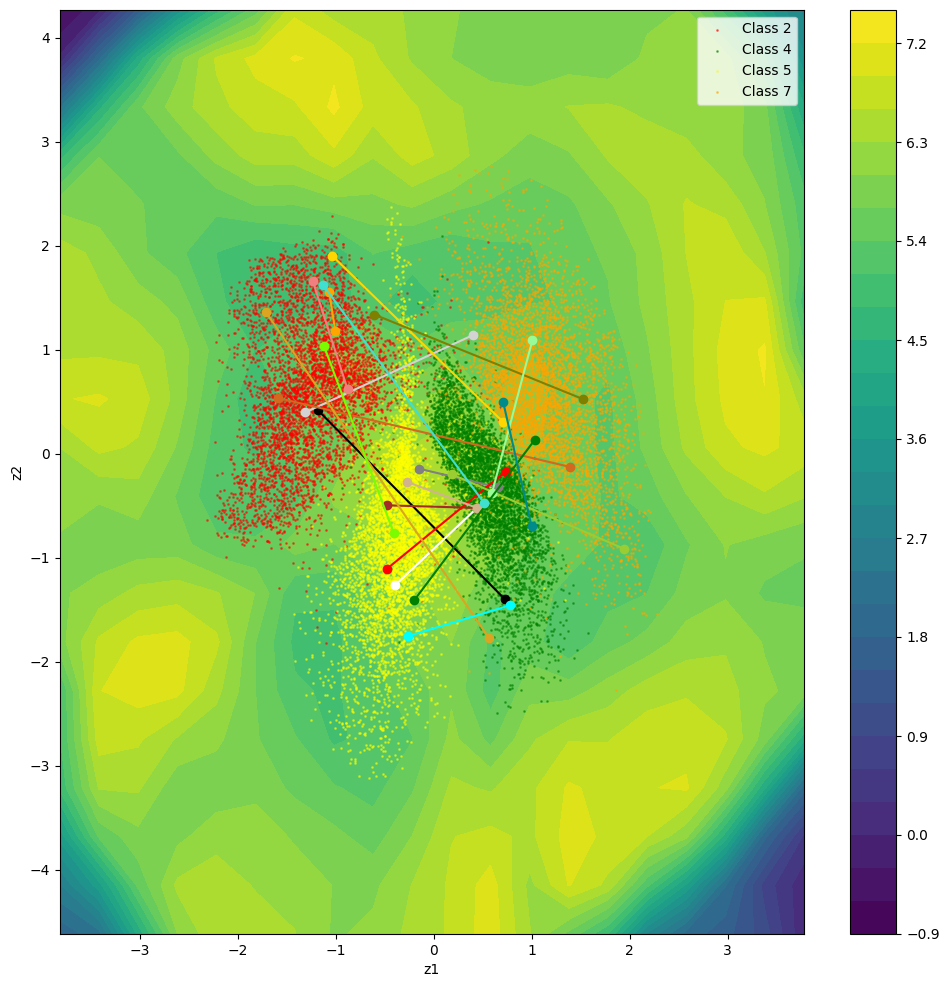}
    \end{subfigure}
\caption{VAE1 can be seen on the left, VAE2 on the right. Both use the improved variance estimate of \citet{arvanitidis2018latentspace}. The colored lines are the 20 interpolation samples, same colors are used in both figures.}
\label{fig:vaelatent}
\end{figure}

\begin{figure}[htb]
    \centering
    \begin{subfigure}{\textwidth}  
        \centering 
        \includegraphics[width=\textwidth]{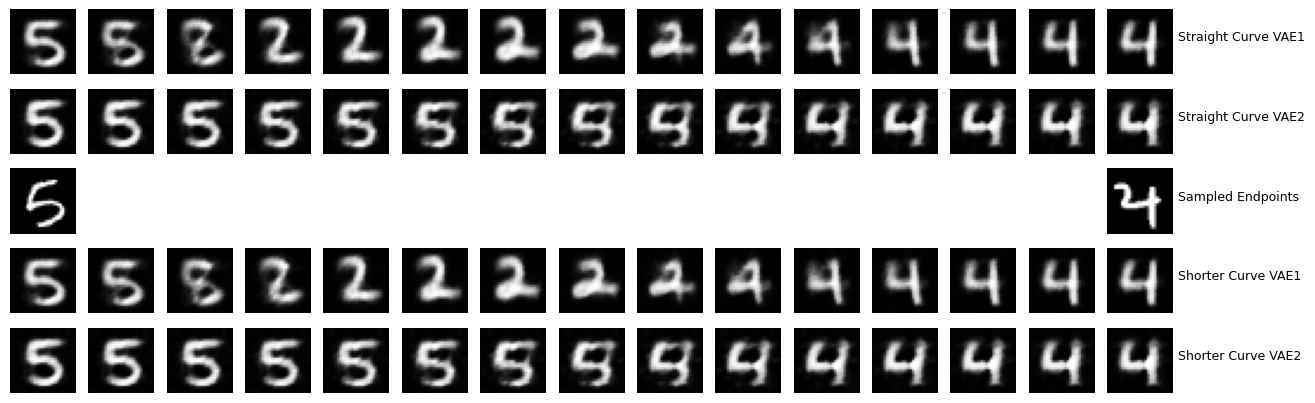}
    \end{subfigure}
    
\caption{Interpolation sequence 19. The original images for start and end point are displayed in the middle row.}
\label{fig:inter}
\end{figure}

We observe that the relative improvements of both models can be vastly different, mainly due to their latent spaces being shaped differently (see Figure \ref{fig:vaelatent}). Therefore interpolations in similar metric neighborhoods should be compared, and from the above samples we could, for instance, choose lines 1, 6, 19 to represent a fair comparison (a fixed threshold should be chosen). One of those interpolations can be seen in Figure \ref{fig:inter}.

\newpage

\section{Feature Mapping to new Output Space: Induced Metric}
\label{app:feature}

Next we derive the effect that chaining a function on a VAE has on the resulting Jacobian/induced metric. We define the decoder of our VAE as: 
\begin{align*}
    \phi (\boldsymbol{z}) = \boldsymbol{\mu}(\boldsymbol{z}) + \boldsymbol{\sigma}(\boldsymbol{z}) \cdot \boldsymbol{\epsilon}
\end{align*}
Where we have $\boldsymbol{z} \in \mathbb{R}^d$, $\phi: \mathbb{R}^d \mapsto \mathbb{R}^D$ and $\boldsymbol{\epsilon} \in \mathcal{N} (\boldsymbol{0}, \mathbb{I}_D)$ (zero-centered multivariate normal distribution with identity matrix covariance).

The result of the decoder of the VAE (with same dimension as input space $X$) then is mapped through some function $h: \mathbb{R}^D \mapsto \mathbb{R}^K$ into the new output space. In total we apply the function $f$ on latent input $\boldsymbol{z}$, with Jacobian:
\begin{align*}
    f(\boldsymbol{z}) &= h(\boldsymbol{\mu}(\boldsymbol{z}) + \boldsymbol{\sigma}(\boldsymbol{z}) \cdot \boldsymbol{\epsilon})
    \\ \\
    \textbf{J}_{Fz} &:= \frac{\partial f(\boldsymbol{z})}{\partial \boldsymbol{z}} 
    = \frac{\partial h(\phi)}{\partial \phi} \frac{\partial \phi(\boldsymbol{z})}{\partial \boldsymbol{z}}
    =: \textbf{J}_{Fx} \textbf{J}_{xz}
\end{align*}

Dimensions: $\textbf{J}_{Fz} \in \mathbb{R}^{K \times d}$, $\textbf{J}_{Fx} \in \mathbb{R}^{K \times D}$, $\textbf{J}_{xz} \in \mathbb{R}^{D \times d}$.
Note: the functions until now should be defined in such a way that the Jacobians are nonsingular.

We follow the proof by \citet{arvanitidis2018latentspace} on the factor $\textbf{J}_{xz}$ by rewriting it as (using element-wise notation of tensors, no einstein summation):
\begin{align*}
    \left( \textbf{J}_{xz} \right)_{ij} &= \frac{\partial f_i}{\partial z_j} 
    \\
    &= \frac{\partial \mu_i}{\partial z_j} + \epsilon_i \frac{\partial \sigma_i}{\partial z_j}
    \\
    &= \left( \textbf{J}_{\mu} \right)_{ij} + \epsilon_i \left( \textbf{J}_{\sigma} \right)_{ij}
\end{align*}
As a result we can split the jacobian $\textbf{J}_{xz} := \textbf{A} + \textbf{B}$.

As mentioned by \citet{arvanitidis2018latentspace}, the resulting ``random'' metric would be $\textbf{M}_{\boldsymbol{z}} = \textbf{J}_{Fz}^T \textbf{J}_{Fz}$, still depending on the normal-distributed variable $\boldsymbol{\epsilon}$. We construct our final metric tensor by taking the expected value of all these ``random'' metric tensors:
\begin{align*}
    \mathbb{E}_{\boldsymbol{\epsilon}} \left[ \textbf{J}_{Fz}^T \textbf{J}_{Fz} \right] &= \mathbb{E}_{\boldsymbol{\epsilon}} \left[ \left(\textbf{J}_{Fx} \textbf{J}_{xz}\right)^T \left(\textbf{J}_{Fx} \textbf{J}_{xz}\right) \right]
    \\
    &= \mathbb{E}_{\boldsymbol{\epsilon}} \left[ \textbf{J}_{xz}^T \textbf{J}_{Fx}^T \textbf{J}_{Fx} \textbf{J}_{xz} \right]
    \\
    &= \mathbb{E}_{\boldsymbol{\epsilon}} \left[ \textbf{J}_{xz}^T \textbf{M}_{Fx} \textbf{J}_{xz} \right]
    \\
    &= \mathbb{E}_{\boldsymbol{\epsilon}} \left[ (\textbf{A} + \textbf{B})^T \textbf{M}_{Fx} (\textbf{A} + \textbf{B}) \right]
    \\
    &= \mathbb{E}_{\boldsymbol{\epsilon}} \left[ \textbf{A}^T \textbf{M}_{Fx} \textbf{A} + \textbf{\textbf{A}}^T \textbf{M}_{Fx} \textbf{B} + \textbf{B}^T \textbf{M}_{Fx} \textbf{A} + \textbf{B}^T \textbf{M}_{Fx} \textbf{B} \right]
    \\
    &= \mathbb{E}_{\boldsymbol{\epsilon}} \left[ \textbf{A}^T \textbf{M}_{Fx} \textbf{A} \right] 
    + \mathbb{E}_{\boldsymbol{\epsilon}} \left[ \textbf{A}^T \textbf{M}_{Fx} \textbf{B} \right] 
    + \mathbb{E}_{\boldsymbol{\epsilon}} \left[ \textbf{B}^T \textbf{M}_{Fx} \textbf{A} \right] 
    + \mathbb{E}_{\boldsymbol{\epsilon}} \left[ \textbf{B}^T \textbf{M}_{Fx} \textbf{B} \right]
    \\
    &= \textbf{A}^T \textbf{M}_{Fx} \textbf{A} 
    + \textbf{A}^T \textbf{M}_{Fx} \mathbb{E}_{\boldsymbol{\epsilon}} \left[ \textbf{B} \right] 
    + \mathbb{E}_{\boldsymbol{\epsilon}} \left[ \textbf{B}^T \right]  \textbf{M}_{Fx} \textbf{A} 
    + \mathbb{E}_{\boldsymbol{\epsilon}} \left[ \textbf{B}^T \textbf{M}_{Fx} \textbf{B} \right]
\end{align*}

In the second to last line we used the linearity of expectation property, and in the last line we used the property that out of all the matrices only \textbf{B} depends on $\boldsymbol{\epsilon}$. Additionally, we know that $\mathbb{E}_{\boldsymbol{\epsilon}} \left[ B_{ij} \right] 
= \mathbb{E}_{\boldsymbol{\epsilon}} \left[  \epsilon_i \frac{\partial \sigma_i}{\partial \boldsymbol{z}_j} \right] 
= \mathbb{E}_{\boldsymbol{\epsilon}} \left[  \epsilon_i \right] \frac{\partial \sigma_i}{\partial \boldsymbol{z}_j} 
= 0$ as the $\boldsymbol{\epsilon}$ is zero-centered. Hence two terms already evaluate to 0 in the expected metric tensor. The last term can be evaluated as follows:

\begin{align*}
    \left( \mathbb{E}_{\boldsymbol{\epsilon}} \left[ \textbf{B}^T \textbf{M}_{Fx} \textbf{B} \right] \right)_{ij} 
    &= \mathbb{E}_{\boldsymbol{\epsilon}} \left[ \sum_{k=1}^D \sum_{l=1}^D B_{ik}^T \: M^{(Fx)}_{kl} \: B_{lj} \right]
    \\
    &= \sum_{k=1}^D \sum_{l=1}^D \mathbb{E}_{\boldsymbol{\epsilon}} \left[ B_{ki} \: M^{(Fx)}_{kl} \: B_{lj} \right]
    \\
    &= \sum_{k=1}^D \sum_{l=1}^D \mathbb{E}_{\boldsymbol{\epsilon}} \left[ \epsilon_k \: \frac{\partial \sigma_k}{\partial \boldsymbol{z}_i} \: M^{(Fx)}_{kl} \: \epsilon_l \: \frac{\partial \sigma_l}{\partial z_j} \right]
    \\
    &= \sum_{k=1}^D \sum_{l=1}^D \mathbb{E}_{\boldsymbol{\epsilon}} \left[ \frac{\partial \sigma_k}{\partial \boldsymbol{z}_i} \: M^{(Fx)}_{kl} \: \frac{\partial \sigma_l}{\partial z_j} \: \epsilon_k \epsilon_l \right]
    \\
    &= \sum_{k=1}^D \sum_{l=1}^D M^{(Fx)}_{kl} \: \frac{\partial \sigma_k}{\partial z_i} \: \frac{\partial \sigma_l}{\partial z_j} \: \mathbb{E}_{\boldsymbol{\epsilon}} \left[ \epsilon_k \epsilon_l \right]
\end{align*}

Here we got rid of all the terms that are independent of $\boldsymbol{\epsilon}$ out of the expected value. What remains is $\mathbb{E}_{\boldsymbol{\epsilon}} \left[ \epsilon_k \epsilon_l \right]$, which we can evaluate from the definition of $\boldsymbol{\epsilon}$, where we know $\text{Cov} \left[ \boldsymbol{\epsilon} \right] = \mathbb{I}_D$:
\begin{align*}
    \left( \text{Cov} \left[ \boldsymbol{\epsilon} \right] \right)_{kl} &= \mathbb{E} \left[ \epsilon_k \epsilon_l \right] - \mathbb{E} \left[ \epsilon_k \right] \mathbb{E} \left[ \epsilon_l \right]
    \\
    &= \mathbb{E} \left[ \epsilon_k \epsilon_l \right] 
    \\
    &:= \delta_{kl}
\end{align*}
Using Kronecker delta $\delta_{kl}$. We once again know that $\boldsymbol{\epsilon}$ is zero-centered, hence has expected value 0. Continuing the computation from above:
\begin{align*}
    \left( \mathbb{E}_{\boldsymbol{\epsilon}} \left[ \textbf{B}^T \textbf{M}_{Fx} \textbf{B} \right] \right)_{ij} 
    &= \sum_{k=1}^D \sum_{l=1}^D M^{(Fx)}_{kl} \: \frac{\partial \sigma_k}{\partial z_i} \: \frac{\partial \sigma_l}{\partial z_j} \: \mathbb{E}_{\boldsymbol{\epsilon}} \left[ \epsilon_k \epsilon_l \right]
    \\
    &= \sum_{k=1}^D \sum_{l=1}^D M^{(Fx)}_{kl} \: \frac{\partial \sigma_k}{\partial z_i} \: \frac{\partial \sigma_l}{\partial z_j} \: \delta_{kl}
    \\
    &= \sum_{k=1}^D  M^{(Fx)}_{kk} \: \frac{\partial \sigma_k}{\partial z_i} \: \frac{\partial \sigma_k}{\partial z_j}
    \\
    &= \left( \textbf{J}_{\sigma}^T \hat{\textbf{M}}_{Fx} \textbf{J}_{\sigma} \right)_{ij}
\end{align*}

As we can see, we only require the diagonal elements of $\textbf{M}_{Fx}$, we denote this diagonal matrix as $\hat{\textbf{M}}_{Fx}$.

Going back to our original equation, we get:
\begin{align*}
    \mathbb{E}_{\boldsymbol{\epsilon}} \left[ \textbf{J}_{Fz}^T \textbf{J}_{Fz} \right] &= \textbf{A}^T \textbf{M}_{Fx} \textbf{A} + \mathbb{E}_{\boldsymbol{\epsilon}} \left[ \textbf{B}^T \textbf{M}_{Fx} \textbf{B} \right]
    \\
    &= \textbf{J}_{\mu}^T \textbf{M}_{Fx} \textbf{J}_{\mu} + \textbf{J}_{\sigma}^T \hat{\textbf{M}}_{Fx} \textbf{J}_{\sigma}
    \\
    &=: \textbf{M}_{Fz}
\end{align*}

Hence when adding mappings on top of the decoder mapping of the VAE, this is how we compute the total induced metric tensor.

\end{document}